\documentclass[twoside]{article}[12pt, a4paper]

\usepackage{graphicx}
\usepackage[a4paper,top=2.5cm,bottom=2.5cm,left=3cm,right=3cm,marginparwidth=1.75cm]{geometry}
\usepackage{authblk}
\usepackage{graphicx} 
\usepackage[dvipsnames]{xcolor}
\usepackage{array}
\usepackage{tabularx}
\usepackage{booktabs}
\usepackage{float}
\usepackage{subcaption}
\usepackage{parskip}
\usepackage{setspace}
\usepackage{url}

\usepackage{fancyhdr}
\newcolumntype{C}[1]{>{\centering\arraybackslash}p{#1}}

\title{Collaborative Language Models for Semantic IoT Data Processing: Validation through Zero-Shot Human Activity Recognition}

\author[1]{Mohsen Shirali}
\author[1]{Kean Ooi}
\author[1]{Anna Knab}
\author[2]{Mohammadreza Fani Sani}
\author[1]{Estefanía Serral}

\affil[1]{Research Centre for Information Systems Engineering (LIRIS), KU Leuven, Brussels 1000, Belgium (\texttt{\{mohsen.shirali,estefania.serralasensio\}@kuleuven.be})}
\affil[2]{Microsoft Development Center Copenhagen, Copenhagen, Denmark (\textit{mfanisani@microsoft.com})}

\date{}

\begin{document}

\maketitle

\small{Corresponding author: Mohsen Shirali (\texttt{mohsen.shirali@kuleuven.be})}

\begin{abstract}
Internet of Things (IoT) systems generate low-level sensor observations that are not directly suitable for process analysis and must first be transformed into semantically meaningful events. This paper investigates collaborative language models as a zero-shot semantic abstraction layer for converting such observations into higher-level event and activity labels. We propose a five-stage workflow comprising semantic mapping, semantic aggregation, temporal and contextual aggregation, specialized reasoning, and final event abstraction. The resulting labels and contextual information can subsequently be represented as event records for downstream Process Mining (PM) tasks. The approach is evaluated through Human Activity Recognition on the CASAS Aruba smart home dataset using eleven workflow configurations. The experiments show that collaboration alone does not guarantee improved performance: general-purpose auxiliary SLMs can degrade recognition, whereas assigning complementary specialist roles leads to better results and allows a smaller final classifier to retain much of the performance of a larger model. The selected workflow achieves a macro F1 of 0.643 in the 1.5-day comparison and 0.567 (weighted F1 0.680) in a one-week evaluation. Although trained ML approaches remain more accurate on this dataset, the results demonstrate a different design point: modular semantic event abstraction using off-the-shelf models without task-specific parameter fine-tuning. The findings indicate that the structure of LM collaboration and task decomposition is more important than simply increasing the number of participating models.

\textbf{Keywords.Internet of Things (IoT) \and Small Language Models (SLMs) \and Collaborative Language Models \and Semantic Sensor Data Processing \and Zero-Shot Prompting \and Human Activity Recognition (HAR).
\vspace{20pt}}

\end{abstract}

\section{Introduction}
\label{sec:introduction}

Internet of Things (IoT) technologies continuously generate low-level observations describing movements, device states, physiological signals, environmental conditions, and interactions. Such observations are not directly suitable for Process Mining (PM), which typically relies on event data represented at a meaningful activity level. Consequently, sensor observations must first be transformed into higher-level semantic events before they can support process analysis~\cite{vanzelst2021event,brzychczy2025sensor}. This event abstraction problem is particularly relevant for IoT systems, where data are heterogeneous, distributed across sensors, and strongly dependent on temporal and contextual information~\cite{shirali2024error}.

Developing IoT-enhanced business processes requires aligning process requirements with the interpretation of low-level data and the integration of devices~\cite{valderas2023iotprocesses}. Bertrand \textit{et al.} describe how sensor observations relate to the events and log entries used in PM~\cite{bertrand2022bridging}. We focus on the semantic step that connects observations to process-relevant activity labels.

Language Models (LMs) provide a promising mechanism for this semantic transformation because they can combine textual representations of observations with contextual and domain knowledge through prompting~\cite{annepaka2025large,han2024review}. Recent studies have applied LMs to IoT event abstraction and integration~\cite{shirali2025llm}, zero-shot recognition of Activities of Daily Living (ADLs)~\cite{civitarese2025zero}, textual representations of smart-home sensor observations~\cite{thukral2025layout}, and semantic alignment of heterogeneous sensing datasets~\cite{yan2025lanhar}. However, most existing IoT and HAR approaches rely on a single LM or combine language-based representations with separately trained downstream models~\cite{cleland2024leveraging,thukral2025layout,yan2025lanhar}. In parallel, research on collaborative LMs has explored pipelines, auxiliary models, routing, task decomposition, and edge--cloud collaboration~\cite{chen2025collaborative,hao2024hybrid,li2026collaborative}. How such collaboration can be structured specifically for semantic abstraction of IoT observations remains less explored.

This work investigates that intersection. Rather than asking one model to interpret an entire sensor sequence and directly produce an activity or event label, the proposed framework decomposes semantic abstraction into five complementary stages: semantic mapping, semantic aggregation, temporal and contextual aggregation, specialized reasoning, and final event abstraction. The resulting semantic labels can constitute the activity component of event records and, together with appropriate timestamps, case/object identifiers, and contextual attributes, support subsequent event log construction and process analysis. Our focus is therefore the semantic abstraction step that precedes PM (the effect of the generated events on downstream process discovery or conformance analysis is outside the scope of the present evaluation).

The objective is not to replace conventional supervised HAR when sufficient labelled data and stable deployment conditions are available. Existing trained approaches achieve higher recognition performance on the Aruba dataset~\cite{Yala2017FeatureExtraction,thukral2025layout}. Instead, we investigate a different design point: whether off-the-shelf LMs can perform semantic abstraction without task-specific parameter fine-tuning, while allowing contextual and domain knowledge to be introduced through prompts and distributed among specialized components. Accordingly, \emph{zero-shot} in this paper refers to the absence of task-specific parameter training or fine-tuning; labelled development observations are still used for workflow design and performance evaluation.

We evaluate the approach through an exploratory Human Activity Recognition (HAR) case study using the publicly available CASAS Aruba smart-home dataset~\cite{Cook2013CASAS}. HAR provides a concrete abstraction problem in which sequences of low-level sensor activations must be mapped to higher-level activity labels. Eleven workflow configurations are compared to investigate semantic representation, temporal context, general-purpose versus specialized collaboration, and the capacity of the final decision model. Accordingly, this study addresses the following research questions:

\begin{itemize}
\item \textbf{RQ1.} How can collaborative LMs be organized to transform low-level IoT observations into higher-level semantic events suitable for subsequent process analysis?
\item \textbf{RQ2.} How do general-purpose and specialized collaboration strategies affect recognition performance relative to configurations without auxiliary predictors or specialists?
\item \textbf{RQ3.} To what extent can off-the-shelf LMs perform the abstraction task without task-specific parameter fine-tuning?
\item \textbf{RQ4.} How do semantic representation and the way temporal context is presented in prompts affect activity recognition?
\end{itemize}

The main contributions are threefold. First, we present a five-stage framework that uses LMs as modular semantic-processing components rather than as a single end-to-end classifier. It separates mapping, aggregation, temporal contextualization, specialized reasoning, and event abstraction so that models communicate through explicit intermediate representations. Second, we compare eleven configurations to examine the roles of those representations, specialist inference, and final classifier capacity. This comparison distinguishes running separate specialists from putting the same activity knowledge directly in the final prompt. Third, we evaluate off-the-shelf models without task-specific parameter fine-tuning, reporting a longer V9 run and considering both the flexibility of this design and its recognition gap relative to trained HAR approaches.

The remainder of the paper reviews related work (Section~\ref{sec:related-work}), presents the framework and its HAR instantiation (Section~\ref{sec:approach}), reports the experiments (Section~\ref{sec:results}), and discusses their implications and limitations (Section~\ref{sec:discussion}) before concluding (Section~\ref{sec:conclusion}).

\section{Background and Related Work}
\label{sec:related-work}
Large Language Models (LLMs) are predominantly transformer-based models pretrained on large text corpora and can perform diverse tasks through prompting and in-context information~\cite{vaswani2017attention,han2024review,annepaka2025large}. Their general-purpose capabilities come with substantial computational, memory, and energy requirements. When these requirements lead to cloud-based inference, deployment also depends on network availability and may introduce communication latency, serving costs, and privacy concerns if sensitive observations leave the local environment~\cite{xu2024ondevice,zhang2025rise,burtsev2024working}.

Small Language Models (SLMs) offer a complementary design point. There is no universal parameter threshold separating SLMs from LLMs: published definitions use upper bounds of roughly 5--10 billion parameters, or define smallness by whether a model can run with practical latency on consumer or edge hardware~\cite{lu2024small,corradini2025state,belcak2025small}. Here we use SLM to denote models compact enough to be considered for resource-constrained deployment, not to imply that they were deployed locally in this study. Improvements in training, compression, and efficient inference have increased the capabilities of compact models~\cite{javaheripi2023phi2,kandala2024tinyllm,corradini2025state}. Their smaller resource requirements make local execution more plausible, but their reasoning and generalization capabilities can be more limited on complex tasks~\cite{wang2025comprehensive,hao2024hybrid}.

\subsection{Collaborative and Hybrid Language-Model Workflows}

Collaborative architectures divide inference responsibilities across models through mechanisms such as pipelines, routing, auxiliary models, model fusion, and knowledge transfer~\cite{chen2025collaborative,li2026collaborative}. In a pipeline, one model's output becomes another's input; routing selects which model to invoke, for example according to uncertainty or resource constraints. LLMCad uses a smaller model to propose candidate tokens and a larger one to verify them~\cite{xu2023llmcad}. Hao \textit{et al.} distribute token-level inference between an edge SLM and a cloud LLM~\cite{hao2024hybrid}. Ensemble SuperICL instead incorporates predictions and confidence scores from multiple SLMs into the prompt of a larger model~\cite{mojarradi2024improving}. Other hybrid approaches investigate uncertainty-aware routing and federated coordination~\cite{solat2025federated}, or jointly optimize model caching and inference offloading~\cite{xu2026joint}. These approaches are largely concerned with NLP inference, computational efficiency, or edge--cloud coordination. Our work instead examines collaboration as a means of \emph{semantic task decomposition}, in which models progressively transform fine-grained IoT observations into higher-level labels.

\subsection{Language Models for IoT Event Abstraction and Human Activity Recognition}

Event abstraction from sensor data is an established challenge for PM. Reviews of event abstraction and sensor-based PM emphasize that low-level observations often need to be aggregated and relabelled into activities before event logs can support meaningful process analysis~\cite{vanzelst2021event,brzychczy2025sensor}. Bertrand \textit{et al.} survey how process discovery techniques have been applied to sensor-derived data from smart spaces, including homes~\cite{Bertrand2023ProcessDiscoverySmartSpaces}. In earlier work, Shirali \textit{et al.} used an LLM to transform heterogeneous IoT observations into higher-level event representations, demonstrating how natural language reasoning can support this abstraction step~\cite{shirali2025llm}. For robotic process automation, Fani Sani \textit{et al.} used LLMs to group and label recorded user interactions, illustrating a related abstraction problem with a different data source~\cite{fanisani2024llmsrpa}.

HAR provides a closely related semantic abstraction use case because sequences of sensor activations must be interpreted as higher-level human activities. Cleland \textit{et al.} investigated fine-tuned and non-fine-tuned LMs for activity recognition in smart environments~\cite{cleland2024leveraging}, while Civitarese \textit{et al.} demonstrated zero- and few-shot recognition of ADLs from textual sensor representations~\cite{civitarese2025zero}. TDOST converts smart home sensor triggers into textual descriptions for layout-agnostic HAR~\cite{thukral2025layout}, and LanHAR employs LLM-generated semantic interpretations to improve cross-dataset activity recognition~\cite{yan2025lanhar}. These studies establish the usefulness of language-based sensor representations, but they predominantly rely on individual LMs or trained downstream components.

The present work connects these research directions by treating semantic abstraction as a collaborative LM workflow. Rather than asking one model to solve the complete sensor-to-activity mapping, we evaluate how semantic representation, temporal aggregation, specialized reasoning, and final abstraction can be distributed across off-the-shelf LMs without task-specific parameter fine-tuning.

\section{Proposed Approach}
\label{sec:approach}

\subsection{Overall Approach and Experimental Design}
\label{subsec:overall-approach}
\begin{figure}[t]
    \centering
    \includegraphics[width=\textwidth]{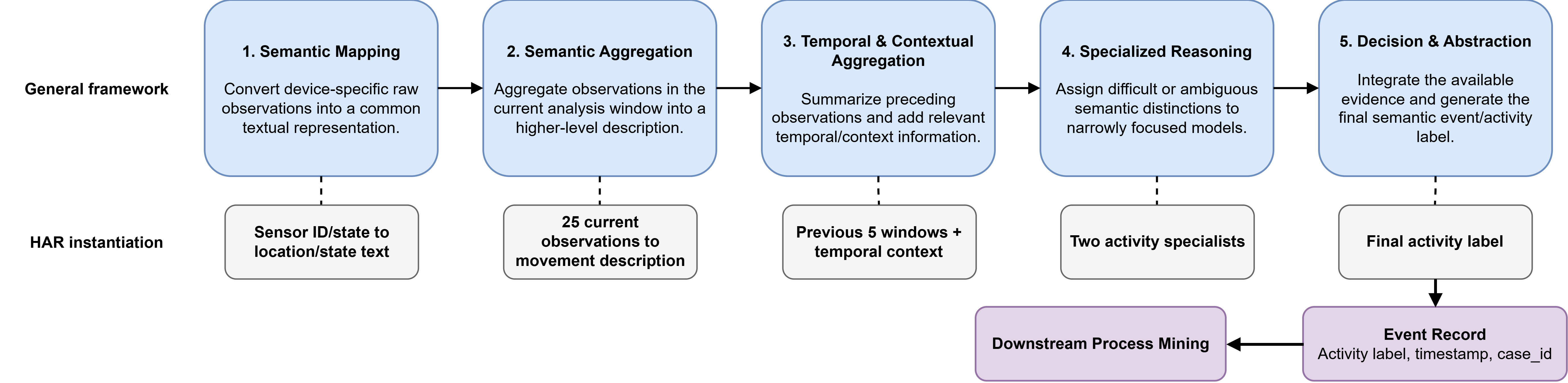}
    \caption{Five-stage collaborative LM framework for semantic IoT event abstraction and its instantiation for HAR. The resulting activity labels can contribute to event records for PM.}
    \label{fig:framework}
\end{figure}

This work investigates whether semantic abstraction of IoT observations can be decomposed into complementary tasks performed collaboratively by multiple LMs. Rather than asking a single model to interpret an entire sequence of low-level observations and directly produce an event label, the proposed framework progressively increases the semantic level through five stages: semantic mapping, semantic aggregation, temporal and contextual aggregation, specialized reasoning, and final decision and abstraction. Figure~\ref{fig:framework} summarizes the general framework and its instantiation for HAR.

The experiments were designed to investigate not only whether collaboration is beneficial, but also how different forms of collaboration affect the result. Eleven configurations vary the effects of sensor-to-language transformation, temporal context, general-purpose and specialized auxiliary models, and the capacity of the final classifier. All LM components are used through prompting without task-specific parameter fine-tuning. Performance is evaluated using accuracy, macro F1, and weighted F1. Each sensor reading inherits the label predicted for its window, and all metrics are computed at the reading level. Table~\ref{tab:activity-results} additionally reports one-vs-rest accuracy and recall for each activity. Macro F1 gives equal importance to each activity, while weighted F1 accounts for class distribution; these metrics are also commonly reported in related HAR studies~\cite{cleland2024leveraging,Yala2017FeatureExtraction}, facilitating contextual comparison.

\subsection{HAR Case Study and Data Preparation}
\label{subsec:case-study}

The framework is evaluated using the CASAS Aruba smart home dataset, which records the daily activities of a single resident through sensors distributed throughout the home~\cite{Cook2013CASAS}. Each observation contains a timestamp, sensor identifier, and sensor state, together with ground-truth ADL annotations. The ground-truth labels are not used to update the parameters of any LM; they are used during workflow development to compare design choices and during evaluation to measure recognition performance. The evaluated activities are \texttt{Meal\_Preparation}, \texttt{Relax}, \texttt{Eating}, \texttt{Work}, \texttt{Sleeping}, \texttt{Wash\_Dishes}, \texttt{Bed\_to\_Toilet}, and \texttt{Housekeeping}.

The inference pipeline uses motion and door-contact observations and skips temperature readings when forming windows. It processes observations chronologically and assigns one activity label to each window. Preliminary development experiments showed that fixed-duration windows produced highly variable numbers of observations, while windows of 50 sensor readings frequently merged short activities with surrounding behavior. We therefore use windows of 25 sensor readings.

The short \texttt{Enter\_Home} and \texttt{Leave\_Home} activities are not target classes at this granularity; neither are \texttt{Respirate} and unlabelled periods. This does not imply that readings from those periods were filtered out. Of 21,622 input readings in the one-week run, 21,574 were matched to a window with a prediction for reading-level scoring. These include 239 readings from non-target activities (91 \texttt{Enter\_Home}, 103 \texttt{Leave\_Home}, and 45 \texttt{Respirate}). Such readings do not count toward ground-truth support for the eight target classes, but they remain in the one-vs-rest denominator and can contribute false positives when assigned a target label.

\subsection{Collaborative Language Models Workflow}
\label{subsec:workflow}

Figure~\ref{fig:framework} defines the five-stage abstraction framework.
In the HAR instantiation, Stage~1 converts each sensor state change into a textual statement describing its mapped location and state while preserving the chronological order of the observations; for example, an activation of a sensor installed in the kitchen is represented as a textual statement indicating movement in the kitchen. Timestamps are subsequently used to derive temporal context. In Stage~2, the Gemma 3n E4B model converts the 25 mapped observations of the current window into a concise natural-language description of the resident's movement. Stage~3 uses the same model to summarize the descriptions of the preceding five windows and supplements them with deterministic temporal information, including the time of day and approximate duration of the current observations.

Stage~4 uses two DeepSeek-R1-0528-Qwen3-8B specialists with reasoning enabled. The specialist roles were defined during workflow development after recurring confusion was observed between \texttt{Relax} and \texttt{Housekeeping}, and between \texttt{Meal\_Preparation} and \texttt{Wash\_Dishes}. The current implementation does not dynamically route individual windows to experts: both specialists are queried for every window. Each receives the current semantic description and contextual information together with activity-specific guidance, and returns one of its assigned activities or ``none'', a confidence estimate, and a short rationale.

In Stage~5, these specialist outputs are not combined through voting or a deterministic conflict-resolution rule. Instead, the final classifier receives the current movement description, temporal summary and context, the five most recent activity predictions, and the structured outputs of both specialists, and produces one activity label. Consequently, conflicting specialist outputs are resolved by the final decision model using the complete context. The selected configuration uses DeepSeek-R1-0528-Qwen3-8B for this stage, whereas the LLM-based configurations use Qwen3 32B with reasoning enabled.

All LM calls were implemented in Python and executed through the OpenRouter API.\footnote{OpenRouter API documentation: \url{https://openrouter.ai/developers}. The implementation, prompts, and experiment outputs are available at \url{https://github.com/mfanisani/collaborative-lms-smarthome-har}.} No Aruba observations were used for parameter training or fine-tuning of the participating models.

\subsection{Workflow Variants and Experimental Protocol}
\label{subsec:variants}
To isolate the contribution of the different processing and collaboration mechanisms, eleven workflow configurations were evaluated. Table~\ref{tab:workflow-variants} summarizes the information supplied to each final classifier. Unless otherwise specified, the LLM classifier is Qwen3 32B with reasoning enabled, while the SLM classifier is DeepSeek-R1-0528-Qwen3-8B.

\begin{table}[t]
\centering
\caption{Evaluated workflow configurations.}
\label{tab:workflow-variants}
\small
\begin{tabular}{C{0.1\textwidth} p{0.6\textwidth} C{0.15\textwidth}}
\hline
\textbf{Version} & \textbf{Input and collaborative components} & \textbf{Final model} \\
\hline
V1  & Raw sensor observations & LLM \\
V2  & Deterministically mapped sensor descriptions & LLM \\
V3  & V2 + SLM-generated history summary & LLM \\
V4  & SLM-generated description of the current window & LLM \\
V5  & V4 + SLM-generated history summary & LLM \\
V6  & V5 + outputs of three general-purpose SLM predictors & LLM \\
V7  & V5 + outputs of two specialized SLMs & LLM \\
V8  & V3 + outputs of two specialized SLMs & LLM \\
\textbf{V9}  & \textbf{V5 + outputs of two specialized SLMs} & \textbf{SLM} \\
V10 & V5 + specialist knowledge placed directly in the prompt & LLM \\
V11 & V5 + specialist knowledge placed directly in the prompt & SLM \\
\hline
\end{tabular}
\end{table}

The three general-purpose predictors in V6 are Mistral-7B-Instruct, DeepHermes-3-Llama-3-8B, and Gemma-2-9B-IT. They receive the same activity recognition task as the final classifier and return a predicted activity, confidence, and short reasoning sentence. V7--V9 instead use the two specialized DeepSeek-R1-0528-Qwen3-8B models described above. V10 and V11 do not run separate specialist models; instead, the activity-specific information used in the specialist prompts is incorporated directly into the final classifier prompt.

Workflow development and evaluation were performed in three phases. First, 1,196 input readings from a morning excerpt on November 6, 2010, were used to develop prompts and select design options such as window size, activity descriptions, and the representation of temporal context. Second, the eleven configurations were compared on 6,165 input readings covering approximately 1.5 days. Each configuration was executed three times (five times for V9 and V11) on the same observations to characterize variation from stochastic LM outputs. The workflow design was iterative: some configurations, particularly the specialist variants, were introduced after observing recurring errors in earlier configurations. The workflow comparison should therefore be interpreted as an exploratory ablation study rather than as an entirely independent model-selection experiment.

Finally, V9 was selected for an extended evaluation on one week of Aruba observations. This experiment was executed once because of computational and API constraints. It provides a broader assessment of the selected configuration, but because the other variants were not rerun over the same week, it does not establish that V9 would remain the best-performing configuration over the longer evaluation period.

\section{Results}
\label{sec:results}
Table~\ref{tab:workflow-results} reports the mean performance of the eleven workflow configurations over three runs (five for V9 and V11) of the 1.5-day evaluation. The results are discussed according to the main design questions: the representation of sensor observations and temporal context, the form of model collaboration, and the capacity of the final decision model.

\subsection{Semantic Representation and Context}
\label{subsec:representation-results}

The representation of low-level observations has a clear effect on recognition performance. As shown in Table~\ref{tab:workflow-results}, transforming raw sensor observations into mapped textual descriptions (V2) improves performance over V1, while adding a summary of preceding windows (V3) provides a further gain. Using an SLM-generated description of the current window (V4) also improves over the mapped representation. However, combining this representation with temporal summarization in V5 does not consistently outperform V3 or V4, suggesting that semantic aggregation and temporal context provide partly overlapping information.

In preliminary prompt-development runs on the morning excerpt, reading-level accuracy was 21.91\% without activity descriptions and 68.56\% in the configuration with them. A prompt with detailed descriptions of up to five preceding windows yielded 51.59\%, compared with 76.51\% in a configuration that used a compressed summary of up to five preceding windows. These were single development runs, not controlled estimates of the effect of individual prompt changes or part of the main workflow comparison.

\subsection{Collaboration Strategy}
\label{subsec:collaboration-results}

The clearest contrast among the collaborative configurations concerns the role assigned to auxiliary models.
V6 performs worse than V5 across all metrics, whereas V7, using specialized experts, achieves the best overall performance (Table~\ref{tab:workflow-results}). The comparison therefore indicates that adding models alone is not sufficient; narrowly defined complementary roles are more useful than redundant general-purpose predictions in the tested configurations.

V7 and V8 use the same specialists and temporal summary but differ in the representation of the current observations. V7 uses the SLM-generated movement description, whereas V8 uses the simpler mapped sensor descriptions. Their macro F1 values are similar (0.659 and 0.653), although V7 achieves higher accuracy and weighted F1. This provides some evidence that the richer semantic representation is beneficial when the observations are subsequently processed by smaller specialist models, although the difference is limited.

The comparisons between V7/V10 and V9/V11 further separate specialist reasoning from the additional activity knowledge contained in the specialist prompts. With the 32B final classifier, V7 and V10 perform relatively similarly, suggesting that the larger model can make effective use of the activity-specific knowledge when it is included directly in its prompt. With the 8B final classifier, however, V9 achieves a macro F1 of 0.643 compared with 0.568 for V11. In this setting, running separate specialist models provides a clearer benefit than providing the same activity-specific knowledge only to the final classifier. The largest activity-level improvement in V9 relative to V11 was for \texttt{Wash\_Dishes}, whose mean F1 increased by 0.361.

\subsection{LLM vs.\ SLM for Final Decision and Abstraction}
\label{subsec:llm-slm-results}

V7 and V9 use the same collaborative structure but differ in the final classifier. Replacing Qwen3 32B with DeepSeek-R1-0528-Qwen3-8B reduces accuracy from 0.827 to 0.794, macro F1 from 0.659 to 0.643, and weighted F1 from 0.838 to 0.807. With only three to five runs per configuration, small differences should be treated as indicative: in two-sample t-tests, the macro F1 difference between V6 and V7 remains significant after Holm--Bonferroni correction, whereas the differences between V7 and V9 are not statistically significant. This does not establish equivalence between the classifiers. V9 was selected for the extended evaluation because it retained much of V7's performance while combining specialist collaboration with a smaller final classifier.

\begin{table}[t]
\centering
\caption{Mean $\pm$ standard deviation of the workflow configurations over three runs (five for V9 and V11).}
\label{tab:workflow-results}
\small
\begin{tabular}{C{0.1\textwidth} C{0.22\textwidth} C{0.22\textwidth} C{0.22\textwidth}}
\hline
\textbf{Version} & \textbf{Accuracy} & \textbf{Macro F1} & \textbf{Weighted F1} \\
\hline
V1  & 0.568 $\pm$ 0.015 & 0.471 $\pm$ 0.012 & 0.556 $\pm$ 0.014 \\
V2  & 0.656 $\pm$ 0.016 & 0.546 $\pm$ 0.004 & 0.677 $\pm$ 0.014 \\
V3  & 0.752 $\pm$ 0.040 & 0.623 $\pm$ 0.045 & 0.764 $\pm$ 0.036 \\
V4  & 0.766 $\pm$ 0.013 & 0.594 $\pm$ 0.032 & 0.774 $\pm$ 0.007 \\
V5  & 0.768 $\pm$ 0.012 & 0.598 $\pm$ 0.033 & 0.774 $\pm$ 0.012 \\
V6  & 0.704 $\pm$ 0.031 & 0.543 $\pm$ 0.013 & 0.711 $\pm$ 0.036 \\
V7  & \textbf{0.827} $\pm$ 0.018 & \textbf{0.659} $\pm$ 0.017 & \textbf{0.838} $\pm$ 0.020 \\
V8  & 0.783 $\pm$ 0.016 & 0.653 $\pm$ 0.065 & 0.790 $\pm$ 0.018 \\
V9  & 0.794 $\pm$ 0.025 & 0.643 $\pm$ 0.025 & 0.807 $\pm$ 0.021 \\
V10 & 0.812 $\pm$ 0.031 & 0.623 $\pm$ 0.048 & 0.815 $\pm$ 0.034 \\
V11 & 0.768 $\pm$ 0.034 & 0.568 $\pm$ 0.027 & 0.769 $\pm$ 0.036 \\
\hline
\end{tabular}
\end{table}

\subsection{Final Workflow Evaluation and Comparison with Existing HAR Approaches}
\label{subsec:extended-results}

The selected V9 configuration was subsequently evaluated on one week of Aruba observations. It achieved an overall accuracy of 0.6615, a macro F1 of 0.5673, and a weighted F1 of 0.6801. These values are lower than those obtained for V9 during the shorter workflow comparison period (0.794, 0.643, and 0.807, respectively).

Table~\ref{tab:activity-results} shows the activity-level results. Performance varies substantially across activities. \texttt{Work} and \texttt{Sleeping} achieve the highest F1 scores, whereas \texttt{Bed\_to\_Toilet} and \texttt{Wash\_Dishes} remain difficult. The low result for \texttt{Bed\_to\_Toilet} is consistent with the limitations of assigning one label to a fixed 25-observation window when an activity is short relative to the surrounding behavior. \texttt{Wash\_Dishes} also remains challenging over the longer period despite the improvement associated with the kitchen specialist in the shorter workflow comparison. For infrequent activities such as \texttt{Bed\_to\_Toilet}, one-vs-rest accuracy is high (0.978) because most readings are true negatives, so recall and F1 are more informative for comparing activities.

\begin{table}[b]
\centering
\caption{Activity-level performance of V9 over one week. Accuracy is one-vs-rest for each activity over all 21,574 evaluated readings; recall is the share of the activity's readings correctly recognised.}
\label{tab:activity-results}
\small
\begin{tabular}{p{0.25\textwidth}C{0.17\textwidth}C{0.17\textwidth}C{0.17\textwidth}}
\hline
\textbf{Activity} & \textbf{Accuracy} & \textbf{Recall} & \textbf{F1} \\
\hline
Work              & 0.9821 & 0.9021 & 0.8657 \\
Housekeeping      & 0.8360 & 0.8078 & 0.6627 \\
Sleeping          & 0.9713 & 0.7518 & 0.8424 \\
Meal\_Preparation & 0.8527 & 0.6555 & 0.7073 \\
Relax             & 0.8817 & 0.5822 & 0.7048 \\
Eating            & 0.9412 & 0.4507 & 0.4782 \\
Bed\_to\_Toilet   & 0.9782 & 0.3708 & 0.1231 \\
Wash\_Dishes      & 0.8788 & 0.2424 & 0.1540 \\
\hline
\end{tabular}
\end{table}

For contextual comparison, Yala \textit{et al.} report an accuracy of 0.8755 and an F1 of 0.6924 on Aruba when excluding unlabelled observations~\cite{Yala2017FeatureExtraction}, while TDOST reports an accuracy of 0.7723 and weighted F1 of 0.7623 in an experiment that includes unlabelled observations~\cite{thukral2025layout}. The proposed workflow therefore does not outperform these trained approaches in recognition accuracy. However, these studies differ in preprocessing, evaluated periods, treatment of unlabelled observations, and training protocols; therefore, the results should not be interpreted as a controlled head-to-head comparison. Their role is to contextualize the recognition performance of the zero-shot collaborative workflow.

\section{Discussion}
\label{sec:discussion}

With respect to \textbf{RQ1}, the experiments show that semantic abstraction can be decomposed into mapping, within-window aggregation, temporal and contextual summarization, specialized reasoning, and final abstraction. The expert-validated IoT--PM bridging model distinguishes this abstraction step from the construction of event log entries~\cite{bertrand2024validatedbridging}. Predicted activities can supply the activity attribute of such entries alongside timestamps and a case notion (e.g., a resident-day). The NICE event log model offers a way to retain links between those entries and their underlying IoT observations~\cite{bertrand2024nice}. We did not construct or evaluate such logs in this study.

Regarding \textbf{RQ2}, the results show that collaboration is beneficial only when model responsibilities are appropriately structured. General-purpose predictors in V6 reduced performance, whereas assigning narrowly defined specialist roles produced better results. The comparisons between V7/V10 and V9/V11 further suggest that separate specialist reasoning is more useful when the final classifier has lower capacity: with the 32B classifier, activity-specific knowledge placed directly in the prompt performs relatively close to explicit specialists, whereas the 8B classifier benefits more clearly from separate specialist reasoning. Similarly, replacing the 32B final classifier in V7 with the 8B model in V9 causes only a modest performance decrease within the tested workflow. Together, these findings indicate that task decomposition and complementary model roles are more important than simply increasing the number or size of participating models.

For \textbf{RQ3}, all participating LMs operate without task-specific parameter training or fine-tuning. Because roles are defined through prompts and textual interfaces, a model could in principle be replaced without retraining the entire workflow, provided the replacement satisfies the role's interface and is evaluated for that role. This is not a quantified saving in training time or computation, nor was development label-free: ground-truth annotations informed choices such as the window configuration and specialist roles.

The results for \textbf{RQ4} show that both semantic representation and temporal context influence recognition quality. Converting sensor identifiers into explicit textual descriptions improves performance over raw observations, while summarizing past windows provides additional contextual information. The preliminary experiments also indicate that compact summaries are preferable to repeatedly supplying detailed historical observations. These findings suggest that prompt construction and information representation should be regarded as part of the workflow design itself, rather than as secondary implementation choices.

An important question raised by the results is why LMs should be considered when established trained HAR approaches achieve higher accuracy on the same dataset~\cite{Yala2017FeatureExtraction,thukral2025layout}. The proposed workflow is not intended to replace supervised classifiers in settings where sufficient labelled data, a stable sensor configuration, and efficient trained models are already available. In such conditions, conventional ML may remain both more accurate and computationally cheaper. The investigated design point is different: \textbf{off-the-shelf LMs provide a semantic interface through which heterogeneous observations, contextual information, and task knowledge can be expressed in natural language and distributed across interchangeable components without task-specific fine-tuning}. Whether this flexibility justifies the additional inference cost depends on the application and cannot be concluded from classification accuracy alone.

\subsection{Limitations and Future Work}

The evaluation uses one smart-home dataset and one resident, and the one-week experiment was executed only once. The November 4--10 evaluation contains both the November 4--5 workflow comparison and the November 6 morning development excerpt, so it is not an independent held-out test. The fixed 25-observation window can disproportionately affect short activities, such as \texttt{Bed\_to\_Toilet}, and cannot represent transitions within a window. Specialist roles were also defined manually after observing recurring classification errors. Beyond these design limitations, noisy sensor streams and gaps in activity annotations can complicate abstraction and assessment, consistent with the data quality issues identified for IoT process mining~\cite{bertrand2023iotdataquality}.

Future evaluations should use independent test periods, longer and repeated runs, adaptive or transition-aware segmentation, and dynamic specialist selection. In another home, the mapping layer could be adapted to different sensor identifiers and locations while preserving the textual interfaces. The PIR-based eSense Smart Home dataset offers a setting for investigating changes in sensor placement~\cite{shirali2023esense}; DAMMI would allow examination of other sensing modalities~\cite{falahrad2024dammi}. In industrial IoT, an analogous workflow could map machine observations to text, summarize their history, and delegate interpretation of specific conditions to specialists before producing process events or machine states. These are architectural propositions, not demonstrated transfer to new homes, residents, or domains.

From a PM perspective, future work should construct event logs from the generated labels and assess their effect on discovery and conformance checking. Context-enriched process discovery from IoT observations~\cite{ahmadi2025contextenriched} and process-mining analyses of variation in daily routines~\cite{difederico2024routinevariation} suggest downstream tasks for evaluating the usefulness of the resulting logs.

The modular design could support future edge or real-time implementations by placing some stages near the sensors and invoking a remote model only when needed. In the tested V9 workflow, a window with prior context requires about five LM calls: one for its description, one for the history summary, two for the specialists, and one for the final decision. Across 808 windows this implies roughly 4,000 calls, a workflow estimate rather than a measured request count. All inference in this study used the OpenRouter API, and latency, energy consumption, communication overhead, privacy, and on-device feasibility were not measured.

\section{Conclusion}
\label{sec:conclusion}

This paper investigated collaborative language models for semantic abstraction of low-level IoT observations and evaluated the approach through a zero-shot smart home HAR case study. The results show that effective collaboration depends on task decomposition rather than model redundancy: general-purpose auxiliary predictors can degrade performance, whereas narrowly defined specialist roles can support difficult semantic distinctions and help a smaller final model retain much of the performance of a larger classifier. The approach does not outperform established trained HAR methods and is therefore better viewed as a complementary design point for modular semantic abstraction without task-specific parameter fine-tuning. By transforming sensor observations into higher-level semantic labels that can contribute to event-log construction, the framework provides a basis for further investigation of language models for IoT event abstraction and downstream PM.

\bibliographystyle{splncs04}
\bibliography{ref}
\end{document}